\documentclass[journal]{IEEEtran}
\ifCLASSINFOpdf
\else
\fi
\usepackage{blindtext}
\usepackage{cite}
\usepackage{amsmath,amssymb,amsfonts}
\usepackage{graphicx}
\usepackage{subfig}
\usepackage{textcomp}
\usepackage{xcolor}
\usepackage{multirow}
\usepackage[T1]{fontenc}
\usepackage[utf8]{inputenc}
\usepackage{mathptmx} 
\usepackage{url}
\usepackage{placeins} 
\usepackage{hyperref}
\usepackage{latexsym}
\usepackage{algorithm,algpseudocode}
\usepackage{makecell}
\usepackage{array}
\usepackage{tabularx}
\usepackage{multirow}
\usepackage{adjustbox}
\usepackage{xcolor}
\usepackage{subfig}

\begin{document}
%
\title{FCN-Pose: A Pruned and Quantized CNN for Robot Pose Estimation for Constrained Devices}
%
%
%

\author{Marrone Silvério Melo Dantas, 
        Iago Richard Rodrigues,
        Assis Tiago Oliveira Filho,
        Gibson Barbosa,
        Daniel Bezerra,
        Djamel F. H. Sadok,
        Judith Kelner,
        Maria Marquezini,
        Ricardo Silva}

\maketitle

\begin{abstract}

IoT devices are a technology recurrent in the contexts of Industry 4.0 and real-time applications. Nonetheless, they suffer from resource limitations, such as processor, RAM, and disc storage. These limitations become more evident when handling demanding applications, such as deep learning, well-known for their heavy computational requirements. A case in point is robot pose estimation, an application that predicts the critical points of the desired image object. One way to mitigate processing and storage problems is compressing that deep learning application. This paper proposes a new CNN for the pose estimation while applying the compression techniques of pruning and quantization to reduce his demands and improve the response time. While the pruning process reduces the total number of parameters required for inference, quantization decreases the precision of the floating-point. We run the approach using a pose estimation task for a robotic arm and compare the results in a high-end device and a constrained device. As metrics, we consider the number of Floating-point Operations Per Second(FLOPS), the total of mathematical computations, the calculation of parameters, the inference time, and the number of video frames processed per second. In addition, we undertake a qualitative evaluation where we compare the output image predicted for each pruned network with the corresponding original one. We reduce the originally proposed network to a $70\%$ pruning rate,  implying an $88.86\%$ reduction in parameters, $94.45\%$ reduction in FLOPS, and for the disc storage, we reduced the requirement in $70\%$ while increasing error by a mere $1\%$. With regard input image processing, this metric increases from $11.71$ FPS to $41.9$ FPS for the Desktop case. When using the constrained device, image processing augmented from $2.86$ FPS to $10.04$ FPS. The higher processing rate of image frames achieved by the proposed approach allows a much shorter response time.

\end{abstract}


\section{Introduction}

Robots had become a common choice when we tried to improve the throughput of many tasks while keeping the reliably \cite{JAVAID202158}. The robot's evolution can lead us to the scenario where we have a complex or dangerous task on the human hand and insert a robot factor into the pipeline, or vice-versa. The systems where a robot and human interact have a set of requirements, and those requirements are the so-called Human-Robot Interaction (HRI) \cite{VILLANI201866}. 

One of the main HRI refers to safety, where we need to ensure that the person interacting with the robot is risk-free, such as collision, burn, and other possible threats \cite{8186877}. Many approaches were developed to ensure safety, notable many of them relaying in some level of intelligence. Those methods, with time, get more and more complex, sometimes requiring an extensive, robust system and response time.  

The HRI places can be significant with many types of equipment or even remote locations where single robots perform risky tasks. Implementing those intelligent solutions can be challenging; first, due to the dimension of the space, we need to ensure a reliable connection between the devices and the processing unity with minimal delay. With the addition of devices, if the processing occurs on a server, the server needs to be robust, increasing the cost of the solutions considerably. One alternative to performing the HRI for safety is deploying it on constrained devices, reducing the network load, and completing the process on the device. Constrained devices are cheap and reliable components, a RaspberryPi, for instance. While they contain limited configurations, such as memory and storage, they can work as edge devices on the network, developing a self-contained solution. 

This work will focus on one of those safety methods, the pose estimation. The pose estimation is a method where we can detect the position of an object inserted in a scenario. Usually, this position is determined by detecting the keypoints that generate the skeleton of the desired object. Those points can return us feedback on the object's position concerning an image o even about the world. For instance, the keypoints can be the joints in a human. With those keypoints, we can work on solutions for HRI methods, such as distance measure, collision detection, and prediction, and intention prediction \cite{coli}. 

The pose estimation is still a current topic of study nowadays; most of the solutions proposed for pose-estimation rely on deep learning, with convolutions neural networks (CNN), for its capabilities and robustness, and is the current state of the art \cite{deepact}. Despite the overwhelming contributions, a downside of CNN is the usual requirement of millions or even billions of parameters that need careful tuning, and this process is highly CPU/GPU demand. Another possible downside in some cases is the disk storage requirements. When CNN's parameters are stored on the disk, they can occupy just a few kilobytes or even gigabytes \cite{radford2019language}. 

For instance, one of its most straightforward architectures, namely LeNet-5 \cite{lenet} with only five convolutional layers and two dense ones, contains around 60 thousand parameters, with 4.81 MB, and the VGG-16, with 138.4 million parameters, on the default setting, can occupy about 528 MB. Furthermore, we observe that even slight modifications to these networks culminate in an exponential increase in the number of parameters. This is the case of AlexNet \cite{alexnet}, where the use of eight convolutional layers and three dense layers has this number of parameters reach as many as 60 million parameters, in other words exhibiting a thousandfold increase.

To execute these robust and resource-demanding solutions in limited environments such as an edge device located or an embedded system is challenging to say the least. To deploy a pose estimation on a constrained, we need to generate an efficient CNN that can output the required keypoints with considerable response time and precision. Reducing the requirements of a CNN is an engaging topic and a highly discussed research area. One of the alternatives is compressing the CNN, where we try to cut the heavy requirements steps of the CNN, being those steps the weights itself, the precision, learning steps, or even another light version of the same CNN \cite{comp_rev}. This paper focuses on two main processes, pruning and quantization. In contrast, the pruning process reduces the total of parameters on the CNN. The quantization reduces the disk storage requirement by lowering the precision of the parameters; for instance, a value that requires 16 bits can be represented as 8 bits. The compression process can increase the response time while keeping a minimal loss o precision. 

In this paper, we propose a solution for pose estimation, where we present a new CNN that can detect the keypoints of a robot, called FCN-Pose. We also tackle the constrained device requirements, applying our proposed model to a pruning technique and quantization. 

Our main contributions are:

\begin{enumerate}
    \item Generate a new convolutional neural network for robot pose-estimation (FCN-Pose) a light compressed version;
    \item Join CNN compression methods of pruning and quantization;
    \item Evaluate our proposed solution on an actual constrained device (Raspberry Pi3)
\end{enumerate}

The rest of this paper is organized as follows. Section \ref{sec:related} shows an overview and literature scan on convolutional neural networks, how the interaction robot-human can occur, and how state of the art handles such subject. It also leverages some methods of reducing the computational demands of CNNs and their affinities with constrained devices. Section \ref{sec:method} shows our image acquisition process and the creation of our scenario for the experiments. At the same time, it provides the details of the generation of the datasets used in our proposed method. Section \ref{sec:proposed} summarizes our proposed solution, first demonstrating our new model, FCN-Pose, the process of pruning and quantization selected, and final, the post-process applied. Section \ref{sec:exp_pose_net} shows our results and validation of the proposed method for our selected scenarios. In this scenario, we also include some discussions related to the results. Finally, Section \ref{sec:conclusion} lists our contributions and future works.

\section{Related Works}
\label{sec:related}

Even with the Convolutional Neural Networks (CNNs) being nowadays state of the art in many fields, such as object detection \cite{Singh2018,Zhang2019}, semantic segmentation \cite{chen2018,Zhang_2019_CVPR}, or image classification \cite{Xie_2020_CVPR,Kolesnikov2019}, they usually are highly resourcing demand (consumption of RAM, CPU, or GPU). Some works are emerging proposing a method for the demand reduction and thus get a fast response time, the so-called CNN compression methods. 

Compressing a CNN can lead the scope of applications to be more extensive, thus embracing real-time applications or augmented reality applications running over embedded systems with limited resources. Two well-known techniques used for compression are pruning and weight quantization.

The pruning process handles the smart removal of a subset of parameters evaluated as less critical for the task. About neural networks, such parameters are filters, layers, or weights. Pruning techniques have been proposed for a long time, with studies dating back to the late 80's \cite{Janowsky1989,23864,80236}. But most recently, due to recent advances in CNN techniques, new pruning strategies may be contemplated. In order to define the parameters that must be pruned, a set of rules were suggested. Lee \cite{lee2019} proposes a global measure for scoring, where sparse layers were pruned after retraining on a spatial domain and a Winograd domain. The work in \cite{Alford2018PrunedAS} considers the filter or layer with the most sparse information as the pruning candidates. Usually, pruning is performed as an offline process, where we already have the trained model, and then we follow the punning operation. But there are works that accomplished the pruning process in parallel with the training. Xiaohan et al. propose an on-the-fly pruning \cite{Xiaohan2019} using a  Stochastic Gradient Descent (SGD) based momentum, zeroing the redundant parameters gradually. 

Besides the punning approaches, other techniques rely on quantization, reducing the required number of bits to represent each weight. Wu et al. \cite{wu2016} used the k-means clustering technique to reduce the representation based on the codebook generated. Another work can be seen as being more straightforward, with the direct precision conversion from the weight to a different bit scale, such as 16 bits or 8bits \cite{37631,Gupta2015DeepLW}. Choukroun et al. \cite{9022167} claim an aggressive precision change, targeting the linear quantization as a mean squared error problem, reaching a 4-bit precision with minimal loss of accuracy.

Pruning and quantization, in general, are orthogonal research areas, but combining them seems a natural path to follow \cite{8824944}. Paupamah et al. \cite{9041096} compare both techniques and show that pruning can help with the overfitting problem despite observing a slight decrease in accuracy on the quantization. On the other hand, the research constructs a smaller and faster neural network. Thung \cite{8578919} proposes an online process of quantization for pruned networks. Here the full-precise network maintains a form of signal for a possible quantization. Ding et al. \cite{8928083} propose an online process; after each epoch, the framework chooses the best technique to proceed with the training and decides whether the next step involves quantization or pruning. Our work aims at one of those weak spots, the shortage of actual experiments on combining pruning and weight quantization, also showing the improvement in the combination. 

Most of the works suggest the need for network compression in order to achieve real-time applications over resource-limited devices. Such devices can be part of the Internet of Things (IoT), sharing information, and processing images and an extensive range of signals \cite{Patel2016InternetOT}. Despite the numerous efforts targeting compression, few of these effectively consider running their solution on devices with limited resources typically used in an IoT or Industry context. For instance, Yao et al. \cite{yao2017} proposes a framework for sensing applications from finding the minimal number of non-redundant hidden elements, and a designed model for mobile devices \cite{Shuochao2017} while evaluating both on an Intel Edison computing platform. Bhattacharya \cite{Bhattacharya2016} proposes an approach to increase the sparsity on CNNs, and evaluates it on four types of limited devices.

Most of the works suggest that we can achieve desirable response time applications over resource-limited devices with CNN's compression. Such devices can be part of the Internet of Things (IoT), sharing information and processing images and an extensive range of signals \cite{Patel2016InternetOT}. For instance, Yao et al. \cite{yao2017} proposes a framework for sensing applications by finding the minimal number of non-redundant hidden elements and a designed model for mobile devices \cite{Shuochao2017} while evaluating both on an Intel Edison computing platform. Bhattacharya \cite{Bhattacharya2016} proposes an approach to increase the sparsity on CNNs, and evaluates it on four types of limited devices. Despite the numerous efforts targeting compression, few of these effectively consider running their solution on devices with limited resources typically used in an IoT or Industry context. We improve this field by adding our experiments and evaluating our proposed combination of pruning and vector quantization on an actual constrained device, a RaspBerryPi 3. 

Our experiments are related to pose-estimation for a robot, in our case, a robot arm, and his relations in a human-robot collaboration (HRC) environment. \cite{6907311} proposes a random decision forest applied on depth images to offer the segments and further the pose estimation. \cite{camera_robot} proposes a framework called Dream (Deep Robot-to-camera Extrinsics for Articulated Manipulators) for pose estimation on robots. This work uses a convolutional neural network, joined with the kinematics of the robotic, to predict the pose estimation based on synthetic data. While having outstanding performance, it requires an expensive setup for real-time performance and does not add human interaction to the process. \cite{6677313} proposes a system for 3D pose estimation for a quadruped robot, mainly based on an IMU sensor, a set of gyro, and accelerometer sensors. In many of those works, we face some limitations. On one side, we can get a good performance but requires additional resources for the decision pipeline. 

Unfourtanaly, just a few works tackle the HRC related to posing detection. For instance, \cite{Rodrigues2022} proposes a safety collision detection but relies on robot and human segmentation for the task. While presenting good results, we can improve the precision of those tasks with pose estimation. Some of those are more just direct guidelines \cite{guidesafety, Bicchi2008}. Improving that area with some intelligence level can further reduce the automatization process's risks and manual load. 

In this study, we consider using a constrained device to perform pose-estimation with a pruned and quantized CNN to handle further problems related to HRI. To the best of our knowledge, no other work jointly combines those requirements, being this one of our main contributions.

\section{Methodology}
\label{sec:method}

In this section, we will explore the process of dataset generation and data augmentation. How the proposed CNN was proposed, he was trained, and his post-processing.

\subsection{Datasets}

This work tries to solve the problem of robot pose detection with segmentation data. Unfortunately, as far as our research goes, there is no public dataset available for robot pose or robot segmentation. We generate two new datasets to overcome this limitation: keypoints for pose detection and segmentation for data support. Our data is the same as used on \cite{silva2020assessing,Gripper-IJAC}, but since the data only has the full segmentation of the robotic arm, we need to make some additions. 

\subsubsection{Data Description}

\begin{figure}[ht!]
    \centering
    \resizebox{0.8\columnwidth}{!}{\includegraphics{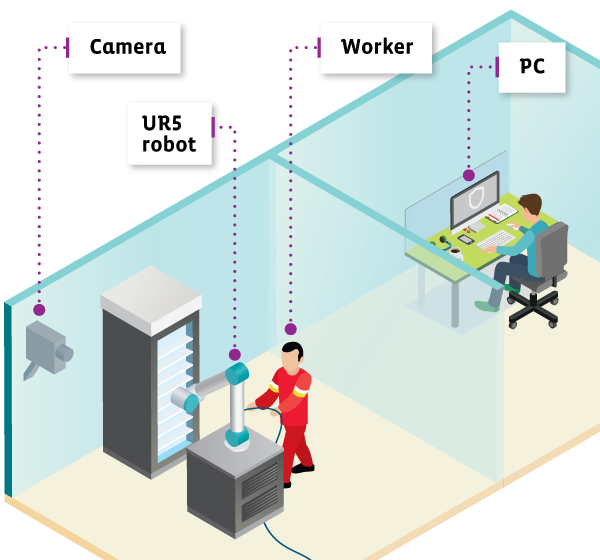}}
    \caption{Scenario overview (same as Silva et. al \cite{silva2020assessing}.}
    \label{fig:scenario_overview}
\end{figure}

The scenario comprises a well-controlled environment simulation. It represents human-robot interaction. A UR5 robotic arm performs maintenance from Universal Robotics with five degrees of freedom. Regarding image capture, a web camera, Logitech C270, recording at 20 frames per second in 1080p. A volunteer assists the robotic arm in-loco. Figure \ref{fig:scenario_overview} shows the scenario´s overview.

\FloatBarrier

The dataset contains approximately 20,000 frames. For the data annotation, we chose to follow a filtering strategy. We select 1000 frames from all the video files, around one frame at each second. We consider that reducing the data also decreases the likelihood of overfitting. The frames of a video generate a strong interdependency; each frame has a slight change, being closely similar to its neighbors, so removing those samples can lead to a diverse dataset and a better generalization. Figure \ref{fig:scenario_full}, depicts a sample frame.

\begin{figure}[ht!]
\begin{center}
\includegraphics[width=0.9\linewidth]{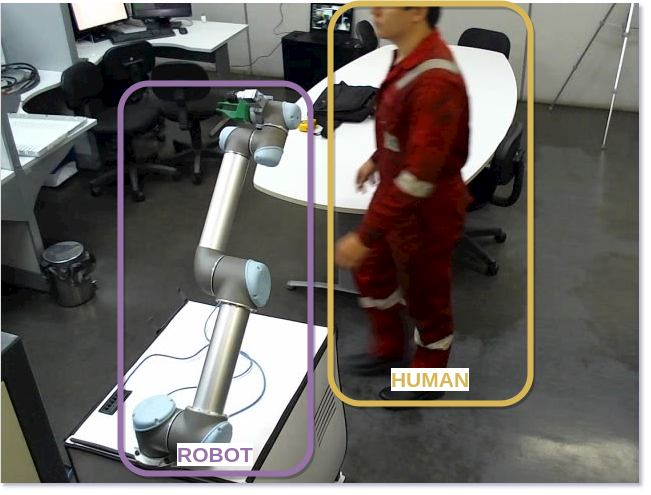}
\caption{Sample of the image acquired from the webcam. On the left, we have the robotic arm; on the right, the human.}
\label{fig:scenario_full}
\end{center}
\end{figure}
\FloatBarrier

\subsubsection{Keypoints Annotation}

As shown in Figure \ref{fig:scenario_full}, one can see the robotic arm and our human volunteer, but this image is raw. We add a plain overlay for better visualization in the sample image, but we first need to generate some manually annotated data for the pose detection estimation training. This data comprises the set of keypoints that compose the robotic arm. We consider each elbow as a keypoint, obtaining a total of 8 keypoints. It is valid to notice that the total of keypoint is bonded to the model of a robotic arm, may in another type of robot, the total of keypoint changes. The VGG Image Annotator (VIA) performed the annotation \cite{dutta2019via}. Fig. \ref{fig:scenario_full} shows an example of how the data was annotated.

\begin{figure}[ht!]
\begin{center}
\includegraphics[width=0.9\linewidth]{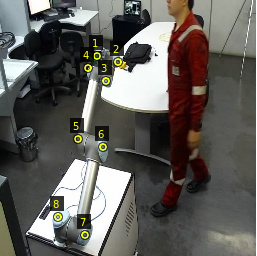}
\caption{Example of robot keypoints annotation with the VIA software.}
\label{fig:scenario_full}
\end{center}
\end{figure}
\FloatBarrier

In cases where the robotic arm presented is most occluded, the positions concerning the camera or the volunteer body were removed. We consider that those cases could increase the noise scope on the dataset.

\subsubsection{Segmentation Annotation}

Concerning our approach, our main target was the keypoints detection but based on the segmentation task. After keypoints annotation, we generated our segmentation dataset. We generated a segmentation mask for each point for each sample image with the eight keypoints annotated. We also generated a segmentation mask referent to the line segmentation that connects the robot skeleton's keypoints.  

\begin{figure}[ht!]
\begin{center}
\includegraphics[width=0.99\linewidth]{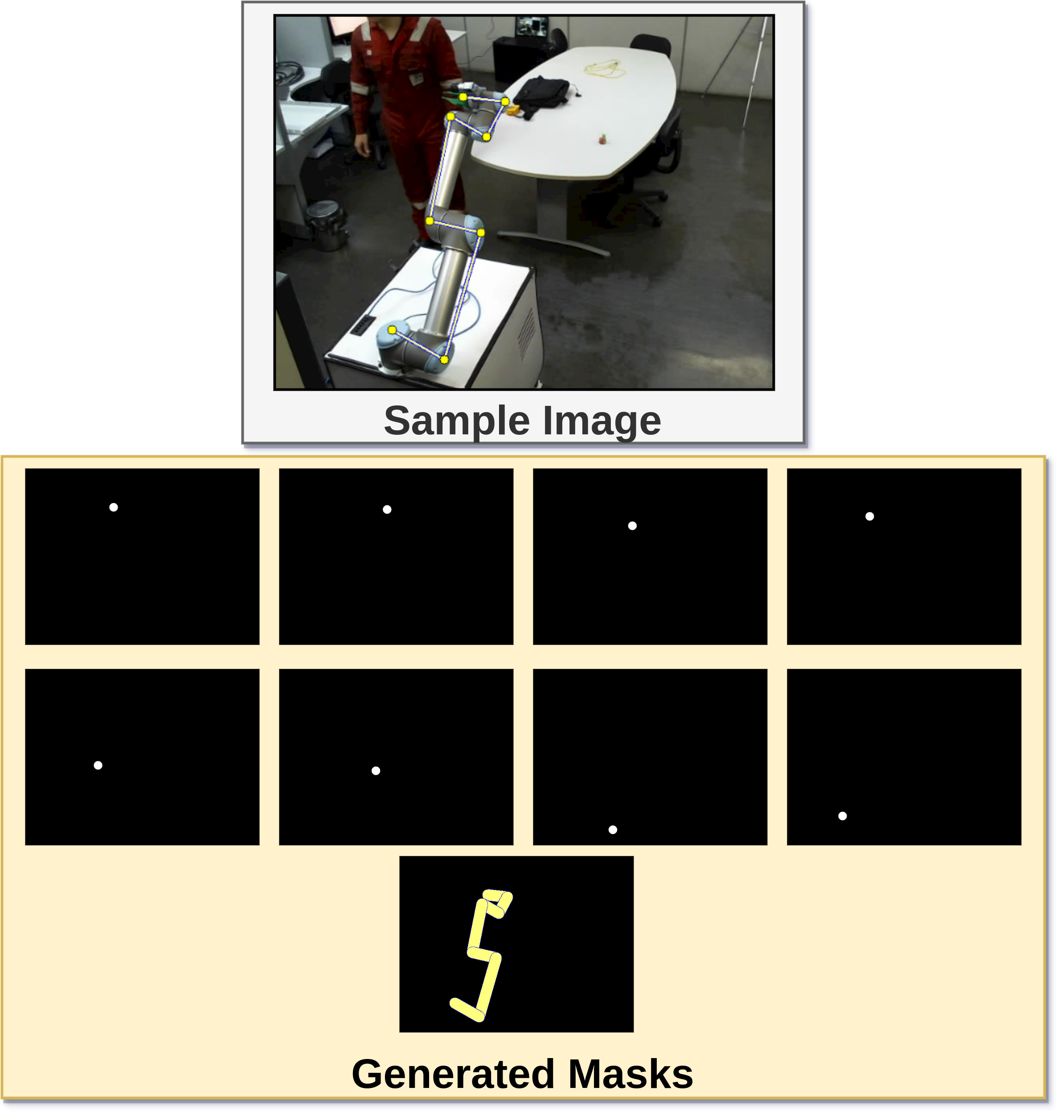}
\caption{Sample of segmentation masks generated. First we have a input image, and as output we generate the set of masks for the keypoints and skeleton.}
\label{fig:segmentation_process}
\end{center}
\end{figure}

Finally, we ended up with nine masks for each sample and a set of annotated keypoints. Eight are related to keypoints, and the last one is to the full robot skeleton. Figure \ref{fig:segmentation_process} shows a sample of the segmentation process.

\FloatBarrier

As seen in Figure \ref{fig:segmentation_process}, we have an input image with the annotated keypoints, and we generated all the masks for each point. The points are the region of interest. We chose to select a circular region around the keypoint, expanding it by a radius of 6 pixels. For the robot skeleton, we follow a similar approach. We generate the connection lines between the keypoints and expand next to the thickness by 30. This thickness was enough to cover the whole robotic arm in our case. 

\subsection{Data Augmentation}

As mentioned earlier, we removed some frames due to the correlation factor of the video stream. Still, after this filtering process, we ended up with few images for the training process. To avoid the problem of over-fitting, we chose to perform some data augmentation. The data augmentation process relies on the generation of new images based on the previous samples of the dataset. For instance, changes in rotation, brightness, perspective, and others, can be considered data augmentation techniques. 
After the filtering process, removing the images with few keypoints available due to extreme occlusion, we obtained 508 images. We split the dataset into two sets to perform the data augmentation, a training and validation set. The training set contains $80\%$ of the data or 406 images, and the validation set includes $20\%$ of the data or, more specifically, 102 images. We generated two random data augmentations for each image on the training dataset, enlarging our dataset into 812 samples. After the union of our original dataset with the data augmented data, we got a total of 1218 samples for training.  

For the data augmentation, we chose two techniques: padding and rotation. Padding dislocates the center of the image for each direction, whereas rotation occurs randomly considering the center of the picture. Both methods are considered due to the simulation of the possible camera displacements in a deployment scenario, with some level of collusion for better generalization. During the process of generation, all the data augmentation was random so that we can have the following augmentations and their combinations:

\begin{itemize}
    \item only rotation;
    \item only padding;
    \item rotation and padding;
\end{itemize}

\begin{figure}[ht!]
\begin{center}
\includegraphics[width=0.9\linewidth]{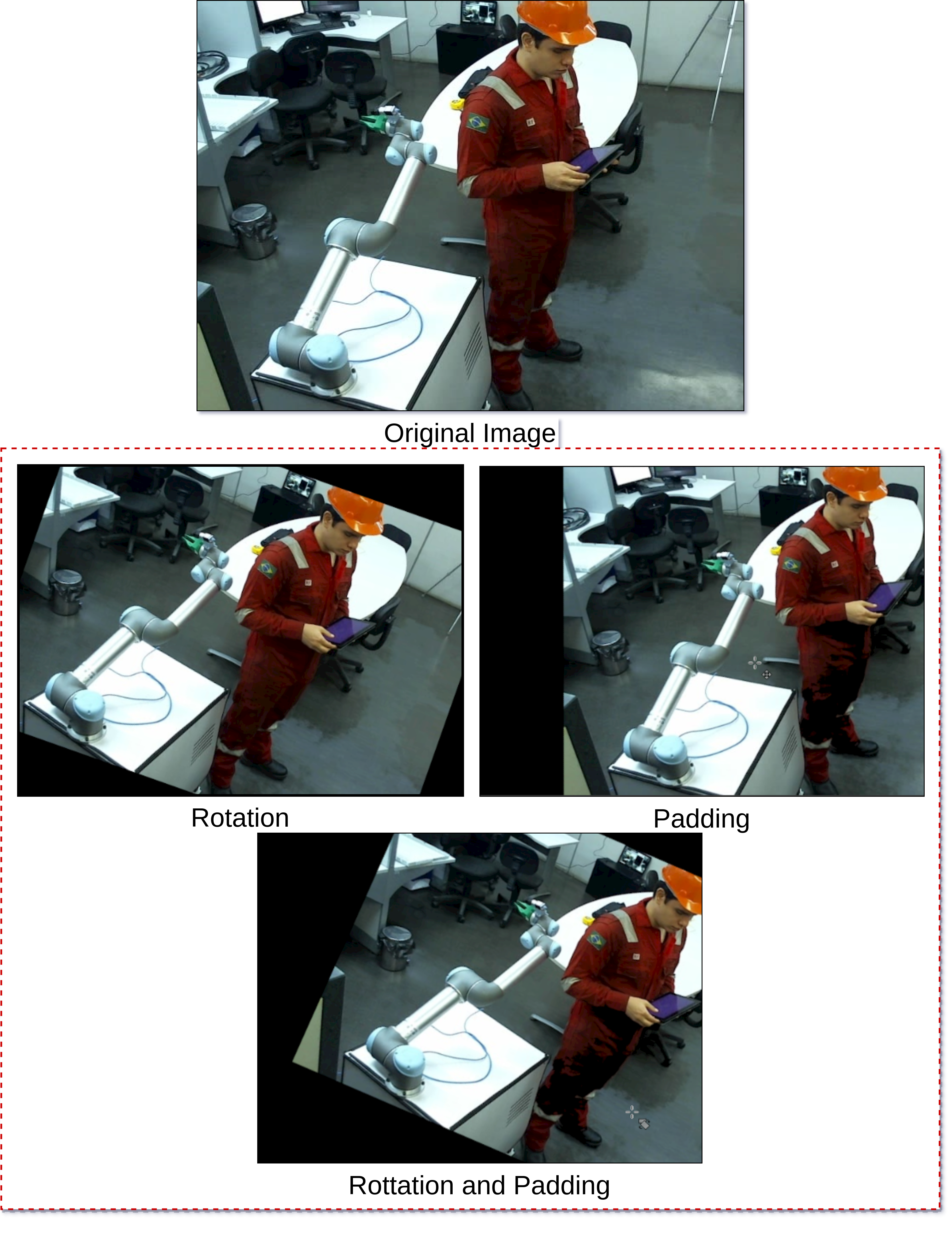}
\caption{Sample of data augmentation. Above we have the original image, and bellow we have the three generated images.}
\label{fig:data_augmentation}
\end{center}
\end{figure}

The images used to generate the data augmentation are also chosen randomly. Figure \ref{fig:data_augmentation} shows a sample of the possible transformations carried out over an image. 

\FloatBarrier

\begin{figure}[ht!]
\begin{center}
\includegraphics[width=0.5\linewidth]{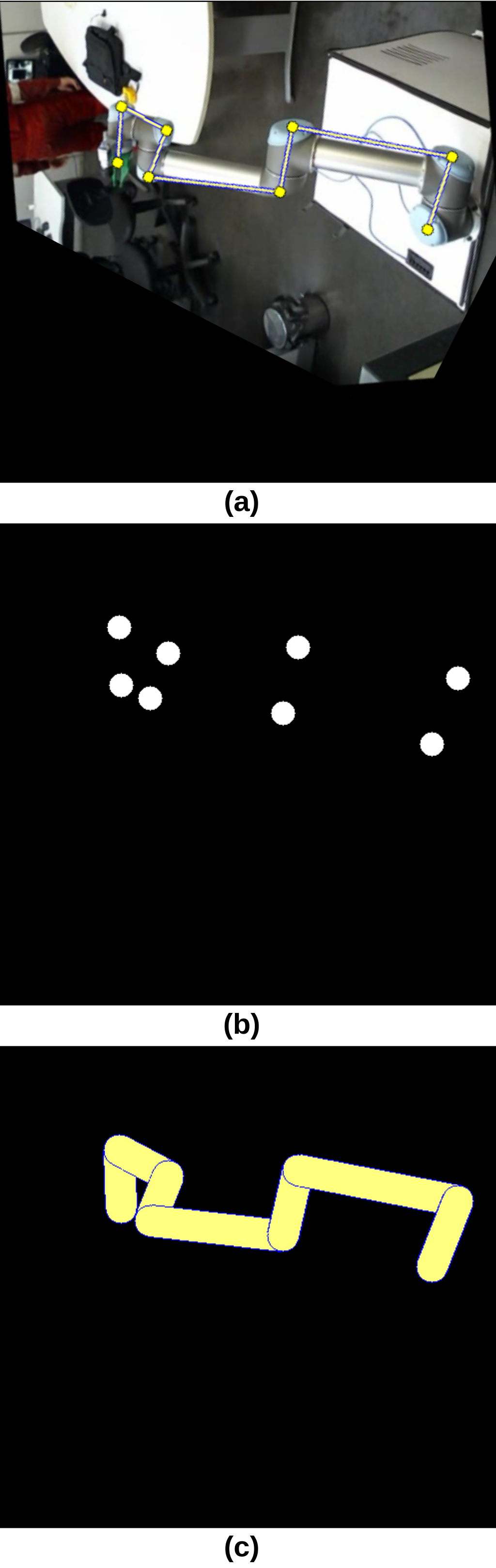}
\caption{Sample of keypoints and masks augmented. In (a) we have the keypoints augmented, (b) the segmented keypoints, here we unified the masks as a single image for better visualization, and finally (c), we have the robotic arm skeleton.}
\label{fig:data_augmentation_sample}
\end{center}
\end{figure}

As illustrated in Figure\ref{fig:data_augmentation}, we generate the images and have some spatial transformation. This leads us to the need to take care of some possible issues. One may lose the robotic arm on the scene with the spatial transformation due to some extreme padding or rotation. We only consider the augmented images with the robotic arm contained in the image. It is valid to notice that the data augmentation was performed with keypoints. Hence, after the augmentation, there is a new set of points. With the new set of points at hand, we also generate the new set of segmentation masks. Figure \ref{fig:data_augmentation_sample} shows a sample of the augmented artifacts.

\FloatBarrier

\section{Proposed Method}
\label{sec:proposed}

This section describes how the Fully Connected Network Pose (FCN-Pose) architecture is built, how the training pipeline worked, and finally presents the additionally performed post-processing refinement.

\subsection{FCN-Pose}

Several well-known architectures for segmentation rely on complex underlying networks, such as U-Net\cite{unet}. Even when considering a lighter version with a smaller backbone, such as MobileNet\cite{mobilenet}, one still needs to handle millions of parameters. The U-Net, and other proposed networks, are indeed powerful but may be considered excessively resource-demanding in the IoT context. Recall that our primary goal is present in two folds: it must first operate close to real-time and secondly execute over devices with limited processing and memory resources. The new model, named FCN-Pose, has been designed with these two requirements in mind.

The FCN-Pose has its architecture heavily based on the work of Jonathan et. al. \cite{segnetfcn}. It nonetheless introduces some major changes. Jonathan proposes an end-to-end semantic segmentation based on Fully Connected Networks; such architecture is an evolution of primitive convents. Previous works on FCN require the definition of a set of parameters. They include size region-based approaches that require the region parameters. They consequently were heavy to process since they needed to classify each region \cite{rcnn}. FCN tries to handle these issues, proposing a network that handles arbitrary input size images and producing a direct output segmentation map. Figure \ref{fig:fcnnetseg} illustrates the FCN model.

\begin{figure}[ht!]
\begin{center}
\includegraphics[width=0.9\linewidth]{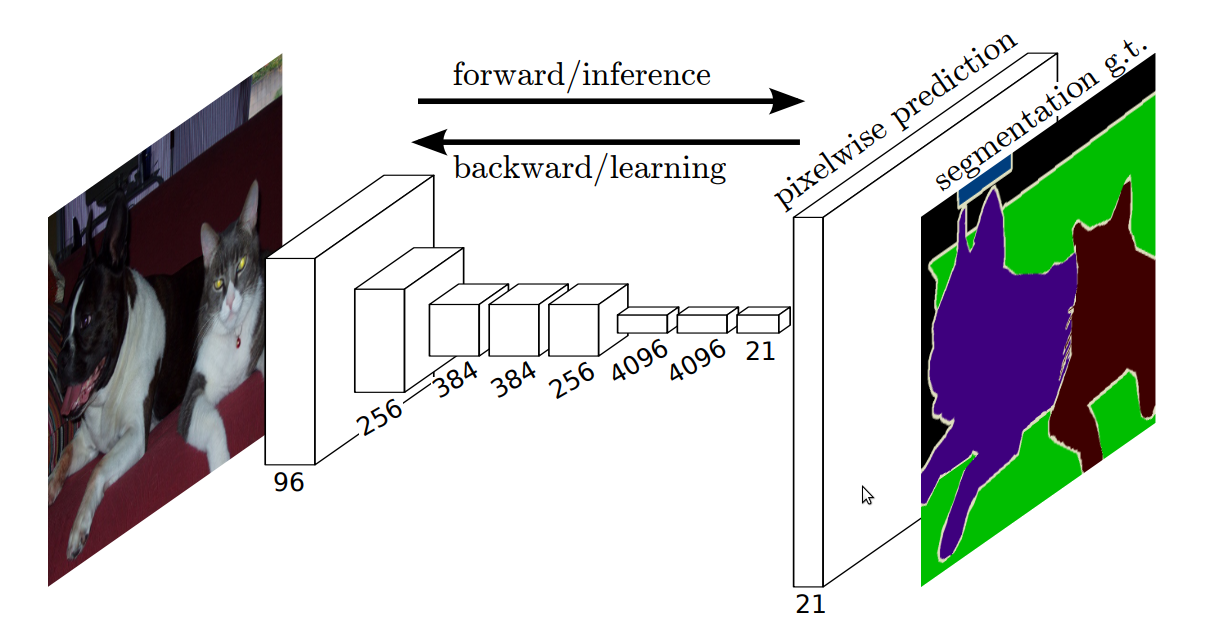}
\caption{Fully Convolutional Networks for Semantic Segmentation. Adapted from \cite{segnetfcn}.}
\label{fig:fcnnetseg}
\end{center}
\end{figure}
\FloatBarrier

As depicted in  Figure \ref{fig:fcnnetseg}, an FCN consists of a stack of convolutional and pooling layers. The stack of layers of the FCN can lead to one major problem: due to the heavy decrease of the feature map, the resolution of the segmentation map is downsampled and can generate some artifacts, such as fuzzy boundaries. To avoid this problem, we extended the FCN with an auto-encoder \cite{autoencoder}, adding a decoder step before the stage of the segmentation map proposal. Figure \ref{fig:fcnnetseg} shows the new proposed overall architecture. 

\begin{figure}[ht!]
\begin{center}
\includegraphics[width=0.9\linewidth]{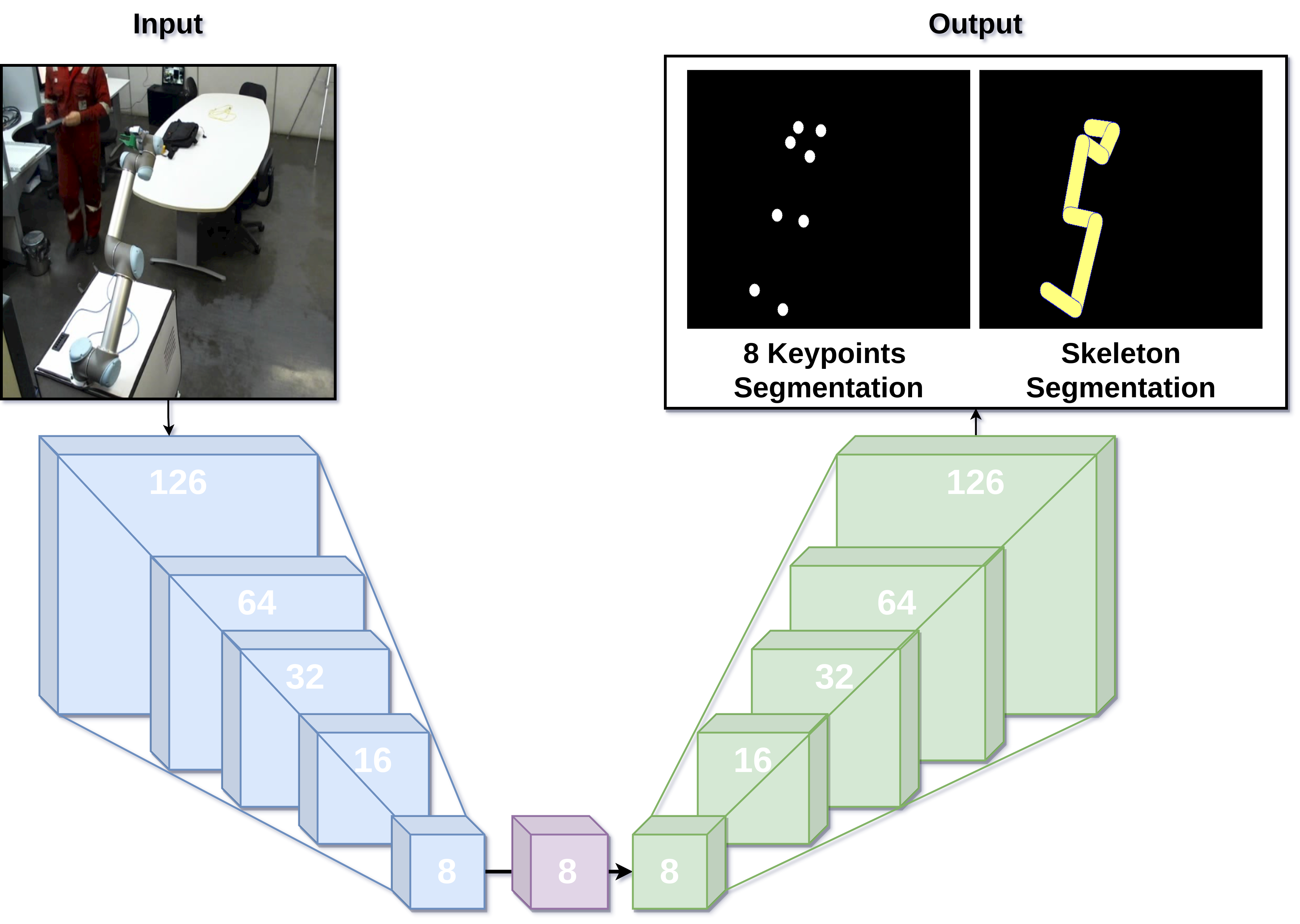}
\caption{FCN-Pose. Structural architecture, with the input on the left, and the proposal with segmentation maps on the right.}
\label{fig:fcnnetseg}
\end{center}
\end{figure}
\FloatBarrier

An examination of Figure \ref{fig:fcnnetseg} shows that our model follows the same proposal for FCN. On the left side in blue, we find an encoding process based on a stack of convolutions and max-pooling layers. There is a decoding process based on a stack of upsampling and convolution layers in green on the right side. The encoding part tries to extract the features from the image and compress them in the latent space (purple part of the image). At the same time, the decoding process aims to improve the map feature reduction and the reconstruction of the segmentation map. 

No additional types of layers were added, such as batch normalization. Besides improving model performance, we sought to keep the model as simple as possible. We believe for our application context and goals that the FCN-Pose model works adequately, despite undergoing a sizeable reduction in the number of its parameters, based on the following observations:

\begin{itemize}
    \item We generate a well-known scenario, with a single target being the robotic arm;
    \item The possible classes to predict are limited;
    \item All our possible predictions correlate. All the keypoints correlate with each other and with the skeleton.
\end{itemize}

An additional segmentation map that was cited before is the skeleton one. We believe that the addition of the skeleton segmentation map, besides the segmentation itself, provides a better spacial generalization for our model. We are adding the skeleton to push the keypoint to follow a structured flow, providing better results, and causing fewer outliers.

The FCN-Pose model consists of 10 convolutional layers, five max-pooling layers, and four up-sampling layers. In all intermediary layers, we use the ReLu activation, whereas we apply the Sigmoid activation for the output layers. In our training, we used input images with 224 pixels of height and 224 pixels of length. Table \ref{tab:model-summary} details the proposed model.

\begin{table}[ht] 
\caption{FCN-Pose Structure} 
\label{tab:model-summary}
\begin{tabular}{lll} 
Model: FCN-Pose \\ \hline 
Layer (type)                   & Output Shape                & Parameters    \\ \hline \hline 
input\_5 (InputLayer)           &  (224, 224, 3)       & 0          \\ \hline 
Convolutional Layer1   & (224, 224, 128)       & 3584       \\ \hline 
MaxPooling Layer 1  & (112, 112, 128)       & 0          \\ \hline 
Convolutional Layer 2   & (112, 112, 64)        & 73792      \\ \hline 
MaxPooling Layer 2  & (56, 56, 64)          & 0          \\ \hline 
Convolutional Layer 3   & (56, 56, 32)          & 18464      \\ \hline 
MaxPooling Layer 3  & (28, 28, 32)          & 0          \\ \hline 
Convolutional Layer 4   & (28, 28, 16)          & 4624       \\ \hline 
MaxPooling Layer 4  & (14, 14, 16)          & 0          \\ \hline 
Convolutional Layer 5   & (14, 14, 8)           & 1160       \\ \hline 
MaxPooling Layer 5  & (7, 7, 8)             & 0          \\ \hline 
Convolutional Layer 6   & (7, 7, 8)             & 584        \\ \hline 
UpSampling Layer 1  & (14, 14, 8)           & 0          \\ \hline 
Convolutional Layer 7   & (14, 14, 16)          & 1168       \\ \hline 
UpSampling Layer 2  & (28, 28, 16)          & 0          \\ \hline 
Convolutional Layer 8   & (28, 28, 32)          & 4640       \\ \hline 
UpSampling Layer 3  & (56, 56, 32)          & 0          \\ \hline 
Convolutional Layer 9   & (56, 56, 64)          & 18496      \\ \hline 
UpSampling Layer 4  & (224, 224, 64)        & 0          \\ \hline 
Convolutional Layer 10  & (224, 224, 9)         & 5193       \\ \hline \hline 
Total parameters: 131,705 \\ \hline 
\end{tabular} 
\end{table}
\FloatBarrier

As observed in Table \ref{tab:model-summary}, we generate as small model with only 131,705 parameters. Note that, for instance, the FCN model, with the VGG16\cite{Simonyan2014}, with the same input dimensions, requires over 130 million parameters. Additionally, our model requires minimal disk storage space. One needs only 1.7 Megabytes to save the full model.

\subsection{Post-Processing}

Once we obtain the activation map, we need to maintain only the relevant region from the keypoints. The generated segmentation can contain some noise levels, with active regions not required. To filter those regions out, we performed a clustering method heavily based on the K-Means clustering method \cite{kmeans}. We first extract each relevant region from the image. Each region will be a cluster. Select the most relevant cluster, and use its centroid as the predicted keypoint. 

While the original algorithm requires setting the clusters (K) total, we do not hold this in information before execution. To overcome this limitation, we added a connectivity restriction. All the points inside a cluster need to be at least a distance $M$, from any other point into the same cluster. Using this restriction, we forced that a region is fully connected within a distance $M$. In our case, we force the $M=1$; hence the region becomes as close as possible. 

We also added self-cluster generation. We generate the initial cluster from an initial point, then start searching for points at a distance $M$, adding them to the cluster. We then repeat the process for each new point. Those points are removed from the initial set of points, then we proceed again to the generation of another cluster, continuing the process until the set is empty. It is valid to notice that, unlike the K-Means, our centroid is not used as an update rule but as a result. At the end of clustering, we generate the centroid, the average of all the points on the cluster. Cluster generation solves our limitation of the K parameters. The Algorithm \ref{alg:expansion} lists the actions of the adopted clustering method. 

\begin{algorithm}
	\caption{Expansion Clustering}
	\begin{algorithmic}[1]
	    \State $S$ = set of points $(x,y)$
	    \State $Clusters$ = set of clusters
	    \State $M$ = Minimum distance
	    
	    \While {$S$ is not empty}
	    
		\For {Each point $p$ in $S$}
		
		\State Create a new cluster $C$
		\State Add the point $p$ in the cluster $C$

           \While {New point is added on $C$}    

		   \For {Each point $p_{i}$ in $C$}
		   
		    \For {Each point $p_{j}$ in $S$}
		    
		        \If {$p_{i}$ != $p_{j}$ and EuclidianDistance($p_{i}$,$p_{j}$)<=M}
		        \State Add the point $p_{j}$ to the cluster $C$
		        \State Remove $p_{j}$ from $S$
		        \EndIf
		    
		    \EndFor

		\EndFor
		
		   \EndWhile
		   \State Add cluster $C$ into $Clusters$
		\EndFor
		
		\EndWhile
		\State Return the centroids of each cluster $C$ in $Clusters$ 
	\end{algorithmic} 
	\label{alg:expansion}
\end{algorithm} 
\FloatBarrier

\begin{figure}[ht!]
\begin{center}
\includegraphics[width=0.4\linewidth]{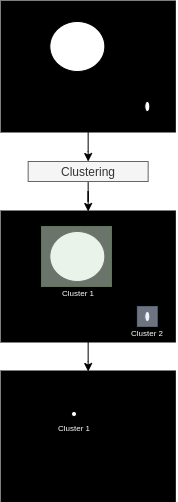}
\caption{Clustering Process. }
\label{fig:clustering}
\end{center}
\end{figure}

The optimal result would be a single cluster with the interest region, but the CNN can produce some noise or artifacts. To solve this problem, we only consider the cluster with the highest number of components in our work and its centroid as our predicted keypoint. Figure \ref{fig:clustering} describes a sample of the workflow for keypoint prediction.

\FloatBarrier

\section{CNN Compressing}
\label{sec:compress}

This section will present how the selected methods for pruning and quantization were applied on the FCN-Pose, and finally, how we proceed with the combination of pruning and quantization. 

\subsection{Pruning}

\begin{figure}[htp]
\begin{center}
\includegraphics[width=0.99\linewidth]{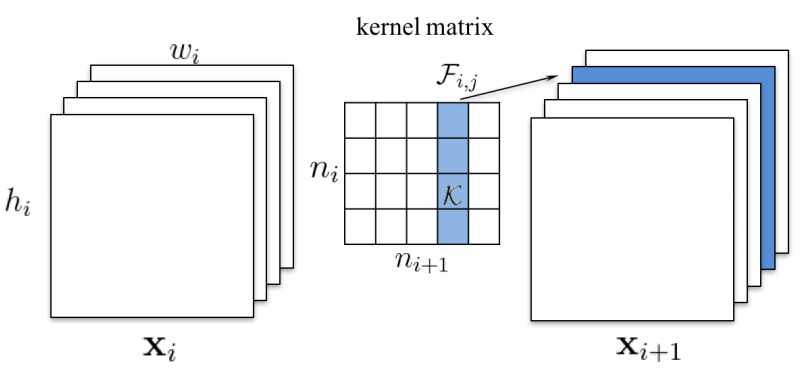}
\caption{Convolution Process. Adapted from Li et al. \cite{Li2017PruningFF}}
\label{fig:conv}
\end{center}
\end{figure}

Our selected pruning approach was the filter ranking pruning technique proposed by Li et al. \cite{Li2017PruningFF}. The pruning filter for each layer is based on the sum of the absolute weights on the filters. Let $n_{i}$ denote the number of input channels for the $i_{th}$ convolutional layer and $h_{i}/w_{i}$ be the height/width of the input feature maps. The convolutional layer transforms the input feature maps $x_{i} \in n_{i}×h_{i}×w_{i}$ into the output feature maps $x_{i}+1 \in R^{n_{i}+1×h_{i}+1×w_{i}+1}$. These are then used as input feature maps for the next convolutional layer. Figure \ref{fig:conv} presents the overall convolution process.

\FloatBarrier

As seen in Figure \ref{fig:conv}, the convolution on a CNN is an attached process, with information shared between the layers and filter, revealing the necessity to propagate the pruning, removing subsequent filters and feature maps relative to the pruned filter. With this sequential pruning, we can reduce the computational cost, getting an $m/n_{i}+1$ of the original computational cost from $m$ pruned filters on a layer $i$. 

\begin{figure}[htp]
\begin{center}
\includegraphics[width=0.35\linewidth]{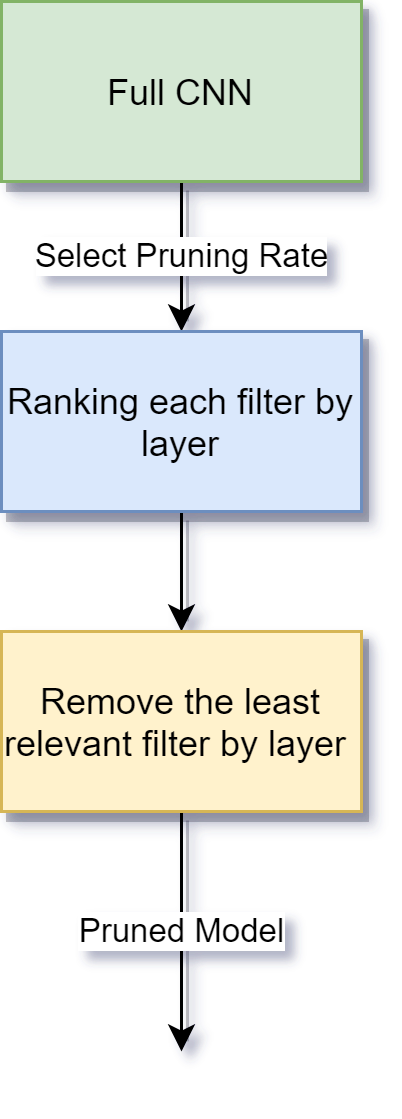}
\caption{Pruning process.}
\label{fig:pruningprocess}
\end{center}
\end{figure}

For the pruning phase, we need to take an indicator of which parameter will be reduced or removed. In our case, we evaluate each filter's importance based on the absolute sum of each filter $\sum \left | F_{i,j} \right |$. As a criterion for data-free selection and as a feature for unbiased selection, $l1-norm$ was chosen. The normalization was required since we have different kernel sizes. Some feature maps naturally can have fewer activation but may represent a relevant feature map. Figure \ref{fig:pruningprocess} shows the pruning pipeline.

\FloatBarrier

This type of pruning worked just fine for us. Since we have a ConvNet, the pruning is straightforward, requiring minimal effort, with no need for feature map projection, and requiring minimal retraining.  

\subsection{Quantization}

We based our quantization on the precision change approaches, converting the single-precision float-point to a half-precision float-point \cite{4610935}. At the same time, usually, the weights are stored as 32 bits (FP32) values. We switch to using a 16 bits values (FP16) standard. Concerning the FP16 representation, there are some slight changes. While the FP32 uses one bit for the sign, 8 bits for the exponent, and finally 23 bits for the significant store, the FP16 standard keeps the one bit for a sign but reduces it to 5 bits for the exponent and 10 bits for the significant precision part. By reducing the float points size from the usual 32 to 16 bits, one can cut almost by half the dimension of the weight while maintaining nearly the same precision \cite{Gupta2015}.

\subsection{Combining Pruning with Quantization}

\begin{figure}[htp]
\begin{center}
\includegraphics[width=0.4\linewidth]{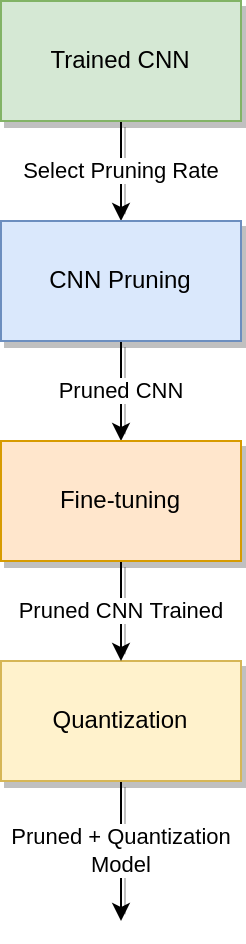}
\caption{Overall compression process.}
\label{fig:flux_general}
\end{center}
\end{figure}

The final step of our compression approach consists of using the pruning process, followed by quantization. The pruning process is the first one applied. Once a pruning rate is selected, the pruning is performed. The second step is related to retraining. During the retraining process, the pruned network is trained again to reduce the degradation caused by the pruning phase. As a final step, we perform post-training quantization, which reduces parameter precision from FP32 to FP16. Figure \ref{fig:flux_general} depicts the entire process.

As seen in Figure  \ref{fig:flux_general}, we used the final CNN without any additional training. The post-quantization phase reduces the time and effort of the process while keeping considerable efficiency and precision. 
\section{Experiments and Results}
\label{sec:exp_pose_net}

This section demonstrates a set of experiments aimed to evaluate if the proposed set o techniques can produce a reliable output in our scenario. By reliability, we mean a model that, even compressed, can result in a slight to no decrease in performance. We also aim to evaluate our model in an actual constrained device to demonstrate the viability of our proposed approach to the scenario for HRI and IoT.

\subsection{Experiment Description}

Concerning the conducted experiments, we first need to set some parameters. The first one is the desired pruning rate. We select a range of values with a slight variation; more specifically, pruning rate variation (or variation step) is set to $10\%$. The $10\%$ steps were chosen based on previous evaluations, we noticed that smaller steps could be repetitive and laborious, representing the tiniest changes, and $10\%$ showed a notable performance in our scenario. As metrics, we collect and observe the 
percentage of correct keypoints (PCK) \cite{9144178}, inference time, the floating-point operation (FLOP), the total number of parameters, the required storage size, and the achieved processed frames per second (FPS).

PCK is an important metric that compares the distance between the predicted and the keypoint. A keypoint is considered correct if the detected keypoint is contained within a distance below a certain threshold value. The distance is calculated pixel-wise. In our case, the distance selected was the euclidian. A $50\%$ distance threshold value is adopted as suggested by many works such as \cite{9144178}. Intuitively, the higher the achieved PCK value, the better our CNN results are. 

The total of parameters represents the learnable values from the CNN. FLOP reflects the computational calculations needed by CNN. Reducing the total number of parameters and FLOP also indicates faster processing and reduced response time. FPS measures the rate of frames we can process per second while still compressing the inference time and the post-processing. Last but not least, we also evaluate the disc storage size needed by the network. It is a critical metric, especially when considering constrained devices that may have limited disc storage space.

Since our work mainly targets limited resource devices, we performed our experiment in two Setups, a Desktop setup, and a constrained device, using a Raspberry Pi 3, the Constrained Device setup. The details of these two setups are listed in Table \ref{fig:setup}. 

\begin{table}[ht]
\centering
\caption{Setup}
\label{fig:setup}
\begin{tabular}{|c|c|c|c|c|c|}
\hline
\textbf{Setup} & \textbf{Processor}                                                                 & \textbf{RAM} & \textbf{S.O.}                                          & \textbf{\begin{tabular}[c]{@{}c@{}}Disc \\ Storage\\ Size\end{tabular}} & \textbf{GPU}                                                                   \\ \hline
Desktop        & \begin{tabular}[c]{@{}c@{}}Intel Core \\ i7-3770 \\ 3.40GHz\end{tabular}           & 16 Gb        & \begin{tabular}[c]{@{}c@{}}Ubuntu\\ 18.04\end{tabular} & 1 TB                                                                    & \begin{tabular}[c]{@{}c@{}}NVIDIA \\ GeForce \\  GTX Titan\\ 6 Gb\end{tabular} \\ \hline
\begin{tabular}[c]{@{}c@{}}Constrained\\ Device\end{tabular}          & \begin{tabular}[c]{@{}c@{}}Quad Core \\ 1.2GHz\\  Broadcom \\ BCM2837\end{tabular} & 1 Gb         & Raspian                                                & 8 GB                                                                    & -                                                                              \\ \hline
\end{tabular}
\end{table}


The GPU was only used for the original training and retraining processes. In the following sections, the predictions running on the Desktop were performed on its CPU. To validate the adopted process, we use K-Fold cross \cite{Refaeilzadeh2009} validation. Data is splinted into K folders. We select a folder for testing at each run, and the remaining data is used for the training. We give K the value 5. For each metric, we obtain its average standard deviation.

\subsubsection{Desktop Setup}

\begin{figure}[htp]
\begin{center}
\includegraphics[width=0.9\linewidth]{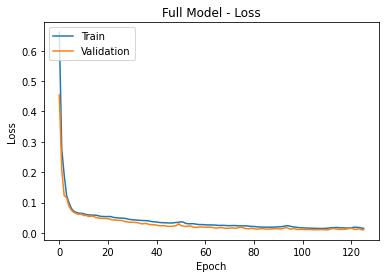}
\caption{Loss plotting in relation from training and validation set.}
\label{fig:training_loss}
\end{center}
\end{figure}

The desktop setup was used in two stages: the training of the base network and the pruning process. The base network represents the full model, in other words, without pruning or quantization. The pruning process also includes the retraining step. For the base training, we followed the pipeline previously mentioned. We trained the FCN-Pose for a maximum of 500 epochs, or until convergence. We used the binary cross-entropy measure to represent a loss in the training phase. Figure \ref{fig:training_loss} shows the average plot of the training models with cross-validation.

Figure \ref{fig:training_loss} shows a considerable decrease at around 20 epochs and stops to learn, converging about 60 epochs. For each folder on the cross-validation, we gather the metrics PCK, inference time, and the estimated FPS. Table  \ref{table:evaluation_fold} offers details of the initial conducted training.

\begin{table}[ht]
\caption{5-Folds Desktop}
\label{table:evaluation_fold}
\centering
\begin{tabular}{|c|c|c|c|}
\hline
\textbf{Folder ID} & \textbf{PCK@0.5} & \textbf{Inference Time} & \textbf{FPS - CPU} \\ \hline
\textbf{0}         & 0.997            & 0.088                   & 11.346             \\ \hline
\textbf{1}         & 0.997            & 0.085                   & 11.731             \\ \hline
\textbf{2}         & 0.999            & 0.084                   & 11.825             \\ \hline
\textbf{3}         & 0.998            & 0.085                   & 11.754             \\ \hline
\textbf{4}         & 0.996            & 0.084                   & 11.899             \\ \hline
\textbf{Average}   & 0.997            & 0.085                   & 11.711             \\ \hline
\textbf{STD}       & 0.001            & 0.001                   & 0.214              \\ \hline
\end{tabular}
\end{table}

\begin{figure}[htp]
\centering
\subfloat[Sample Frame]{%
  \includegraphics[clip,width=0.7\columnwidth]{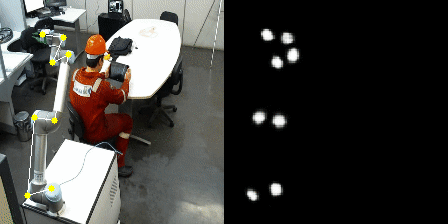}%
}

\subfloat[Sample Frame]{%
  \includegraphics[clip,width=0.7\columnwidth]{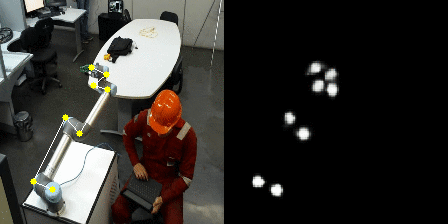}%
}

\caption{Sample Frame. We have the keypoint on the left of the input image, and on the right are the segmentation maps. }
\label{fig:samples_prediction}
\end{figure}

Table \ref{table:evaluation_fold} indicates some good results with an average $PCK@0.5$ of $0.997$. FPS was estimated based on the inference time, being [$1/inference \ time$]. Considering the FPS achieved on the CPU, it can be seen as being real-time, reaching an average of 11.711 FPS. Observe the small standard deviations for PCK and FPS at around $0.001$ and $0.214$, respectively. Figure \ref{fig:samples_prediction} displays some samples of the predicted images. 

\begin{figure}[htp]
\begin{center}
\includegraphics[width=0.7\linewidth]{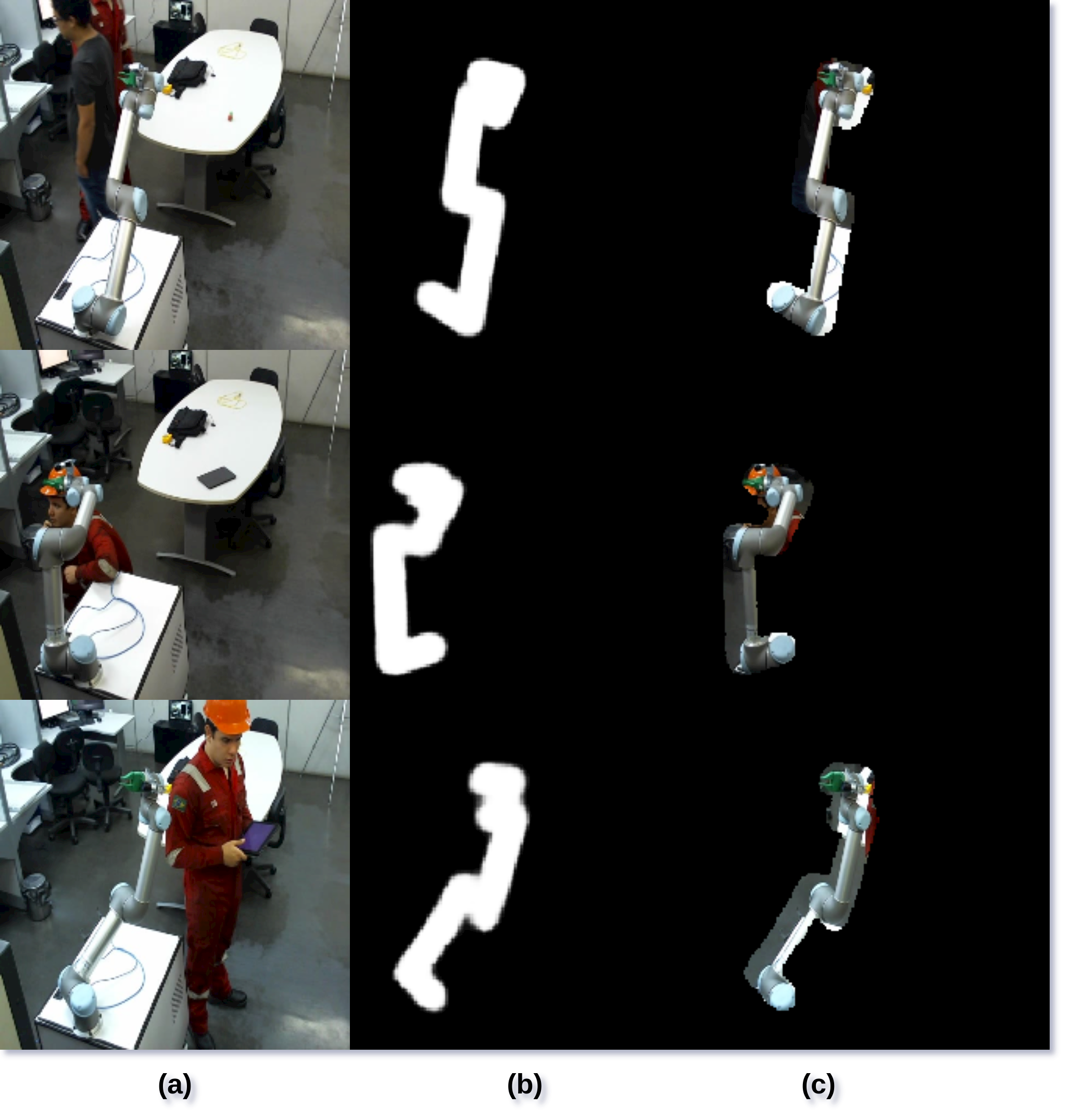}
\caption{Generate segmentation mask for the robotic arm skeleton. First we have the input image (a), the segmentation mask (b), and finally the robotic arm segmented.}
\label{fig:segmentation_pred}
\end{center}
\end{figure}

The left side of Figure \ref{fig:samples_prediction} displays the input image with the predicted keypoints. On the right side, one can see the set of predicted segmentation masks. The segmentation masks were joined to ease visualization. For each point, a predicted mask was joined. The mask referent to the skeleton is hidden. Figure \ref{fig:segmentation_pred} shows only the skeleton segmentation.

Observe that there is in Figure \ref{fig:segmentation_pred} a considerable segmentation mask of the robotic arm. Even in cases where the robotic arm presents some artifacts, the segmentation can focus only on the robotic arm. Our next step is to optimize the network inference time achieved by the pruning process.

We keep the same folds used for the training stage for the pruning process. We alter the pruning rate at a $10\%$ step for each one. We calculate the average and standard deviation values for all metrics. After any pruning, we retrained the CNN for 100 epochs or until the convergence. Table \ref{table:pruning_results} lists the results obtained for the pruning in all rates and the metrics for the full model.

\begin{table}[ht]
\centering
\caption{Pruning Results}
\label{table:pruning_results}
\begin{tabular}{c|c|c|c|c|c|}
\cline{2-6}
\multicolumn{1}{l|}{}                       & \multicolumn{2}{c|}{\textbf{PCK@0.5}}                & \multicolumn{2}{l|}{\textbf{Inference Time}}         & \multicolumn{1}{l|}{} \\ \hline
\multicolumn{1}{|c|}{\textbf{Pruning Rate}} & \textbf{Average} & \multicolumn{1}{l|}{\textbf{STD}} & \textbf{Average} & \multicolumn{1}{l|}{\textbf{STD}} & \textbf{FPS - CPU}    \\ \hline
\multicolumn{1}{|c|}{\textbf{Full Model}}   & 0.997            & \multicolumn{1}{l|}{0.001}        & 0.085            & \multicolumn{1}{l|}{0.001}        & 11.711                \\ \hline
\multicolumn{1}{|c|}{\textbf{10}}           & 0.998            & 0.004                             & 0.089            & 0.002                             & 11.128                \\ \hline
\multicolumn{1}{|c|}{\textbf{20}}           & 0.995            & 0.027                             & 0.070            & 0.001                             & 14.130                \\ \hline
\multicolumn{1}{|c|}{\textbf{30}}           & 0.991            & 0.036                             & 0.072            & 0.001                             & 13.716                \\ \hline
\multicolumn{1}{|c|}{\textbf{40}}           & 0.991            & 0.004                             & 0.062            & 0.000                             & 16.057                \\ \hline
\multicolumn{1}{|c|}{\textbf{50}}           & 0.985            & 0.007                             & 0.035            & 0.001                             & 27.823                \\ \hline
\multicolumn{1}{|c|}{\textbf{60}}           & 0.980            & 0.012                             & 0.034            & 0.000                             & 29.251                \\ \hline
\multicolumn{1}{|c|}{\textbf{70}}           & \textbf{0.980}   & \textbf{0.009}                    & \textbf{0.023}   & \textbf{0.001}                    & \textbf{41.900}       \\ \hline
\multicolumn{1}{|c|}{\textbf{80}}           & 0.912            & 0.051                             & 0.022            & 0.001                             & 44.949                \\ \hline
\multicolumn{1}{|c|}{\textbf{90}}           & 0.284            & 0.092                             & 0.016            & 0.002                             & 61.892                \\ \hline
\end{tabular}
\end{table}

\begin{figure}[htp]
\begin{center}
\includegraphics[width=0.90\linewidth]{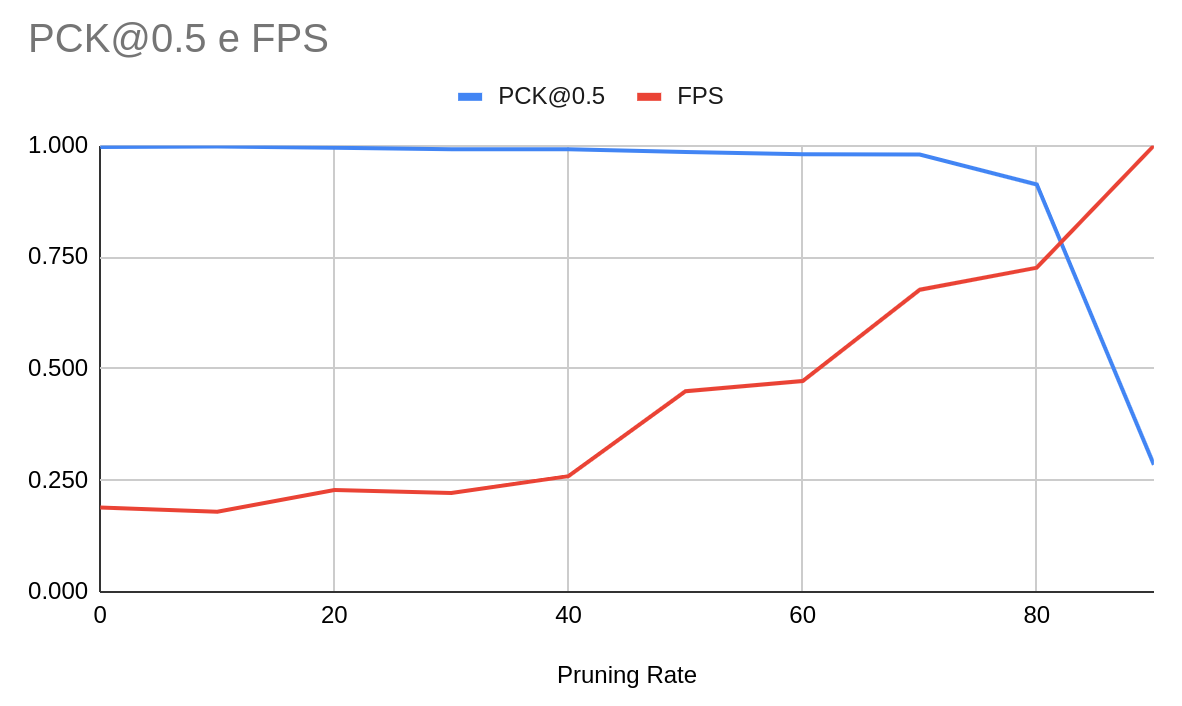}
\caption{PCK versus FPS. The FPS was normalized for a better visualization.}
\label{fig:fps_pck_desktop}
\end{center}
\end{figure}

Table \ref{table:pruning_results} demonstrates a successful pruning process. Observe that up to the $70\%$ pruning rate, and the maximum PCK loss was approximately 0.05 perceptual points (p.p.). After $70\%$ of pruning, CNN suffers a massive degradation. FPS increases largely, from $11.711$ to $41.900$, becoming approximately $3.6$ times faster. Figure \ref{fig:fps_pck_desktop} shows the efficiency plot, of the two metrics PCK versus FPS.

Figure \ref{fig:fps_pck_desktop} presents a sweet spot that comes innately starting from the $70\%$ mark and increases until $90\%$; then we observe a severe degradation. Besides the response time, other important metrics are the storage requirement, the total number of parameters, and finally, the FLOPS, representing the capacity of computational calculation required by the network. Table \ref{table:req_pruning} lists the collected results. 

\begin{table}[ht]
\centering
\begin{tabular}{|c|c|c|c|c|}
\hline
\textbf{\begin{tabular}[c]{@{}c@{}}Pruning \\ Rate (\%)\end{tabular}} & \multirow{2}{*}{\textbf{FPS}} & \multirow{2}{*}{\textbf{Flops}} & \multirow{2}{*}{\textbf{\begin{tabular}[c]{@{}c@{}}Size\\ MB\end{tabular}}} & \multirow{2}{*}{\textbf{Parameters}} \\ \cline{1-1}
\multicolumn{1}{|l|}{}                                                &                               &                                 &                                                                             &                                      \\ \hline
\textbf{\begin{tabular}[c]{@{}c@{}}Full \\ Model\end{tabular}}        & 11.711                        & 262674                          & 1.702                                                                       & 131705                               \\ \hline
\textbf{10}                                                           & 11.129                        & 109909                          & 1.343                                                                       & 110237                               \\ \hline
\textbf{20}                                                           & 14.131                        & 88084                           & 1.092                                                                       & 88375                                \\ \hline
\textbf{30}                                                           & 13.717                        & 67762                           & 0.858                                                                       & 68014                                \\ \hline
\textbf{40}                                                           & 16.057                        & 51049                           & 0.668                                                                       & 51264                                \\ \hline
\textbf{50}                                                           & 27.823                        & 35011                           & 0.482                                                                       & 35185                                \\ \hline
\textbf{60}                                                           & 29.252                        & 24067                           & 0.358                                                                       & 24209                                \\ \hline
\textbf{70}                                                                    & \textbf{41.901}               & \textbf{14563}                  & \textbf{0.248}                                                              & \textbf{14668}                       \\ \hline
\textbf{80}                                                           & 44.950                        & 7138                            & 0.162                                                                       & 7206                                 \\ \hline
\textbf{90}                                                           & 61.893                        & 2449                            & 0.108                                                                       & 2480                                 \\ \hline
\end{tabular}
\caption{Requirements CPU X Pruning Rate.}
\label{table:req_pruning}
\end{table}

\begin{figure}[htp]
\begin{center}
\includegraphics[width=1.01\linewidth]{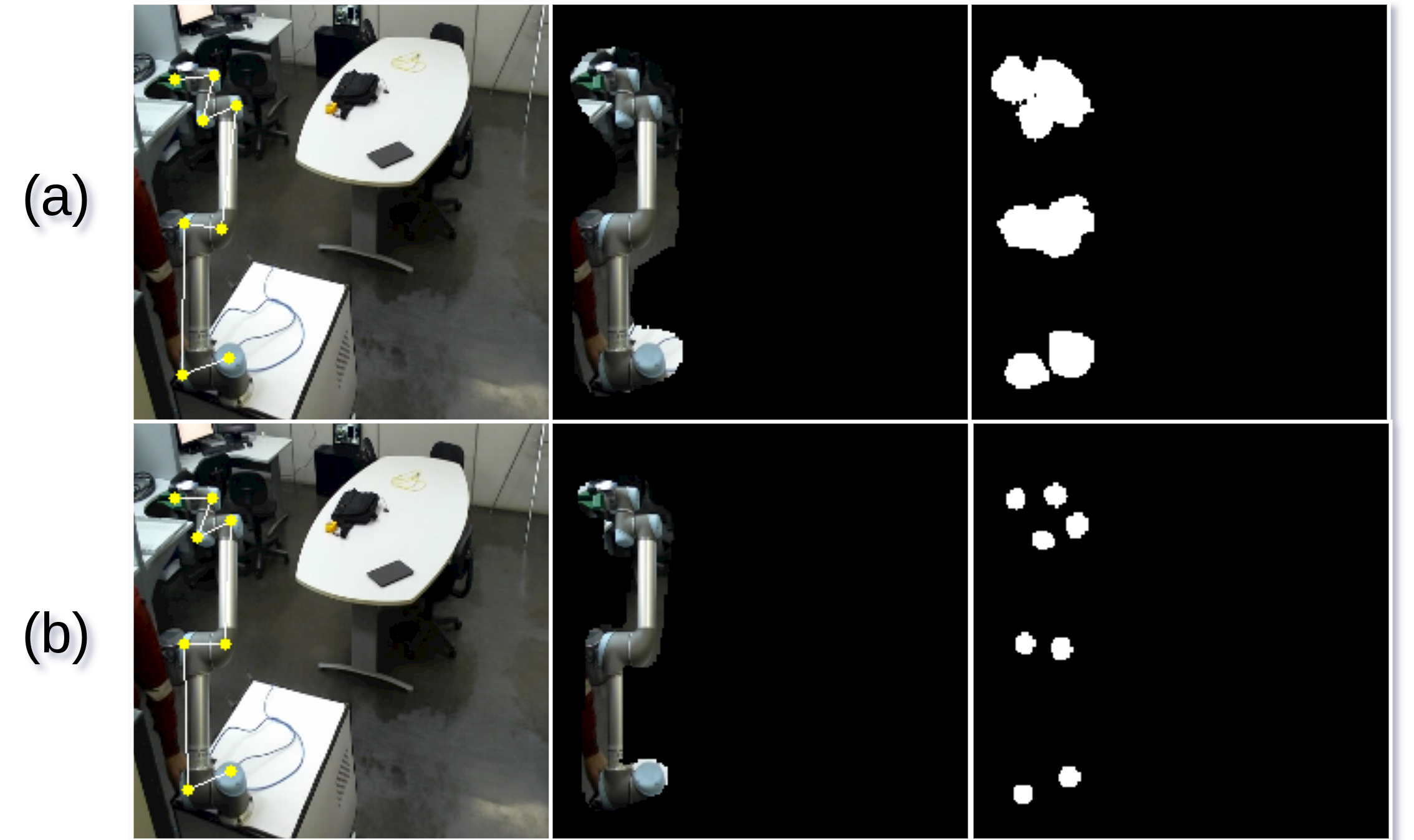}
\caption{Robotic arm and keypoint segmentation. 
We have the robotic arm keypoints on the left column, on the center, we have the robotic arm segmented, and finally, on the right, the keypoints.}
\label{fig:comp_prune_a}
\end{center}
\end{figure}

A close examination of Table \ref{fig:fps_pck_desktop} shows that the total number of the network parameters fell from $131705$ to $14667$, corresponding to an $88.86\%$ reduction. Similarly, the storage requirement fell from $1.702$ to $0.248$, showing an $85.42\%$ reduction. Processing requirements, reflected by the metric FLOPs, also decreased from $262674$ to $12563$, achieving a $94.45\%$ reduction. For better comprehension, we plot some artifacts on the entire process, and figure \ref{fig:comp_prune_a} shows its significant steps. 

\begin{figure}[htp]
\begin{center}
\includegraphics[width=0.7\linewidth]{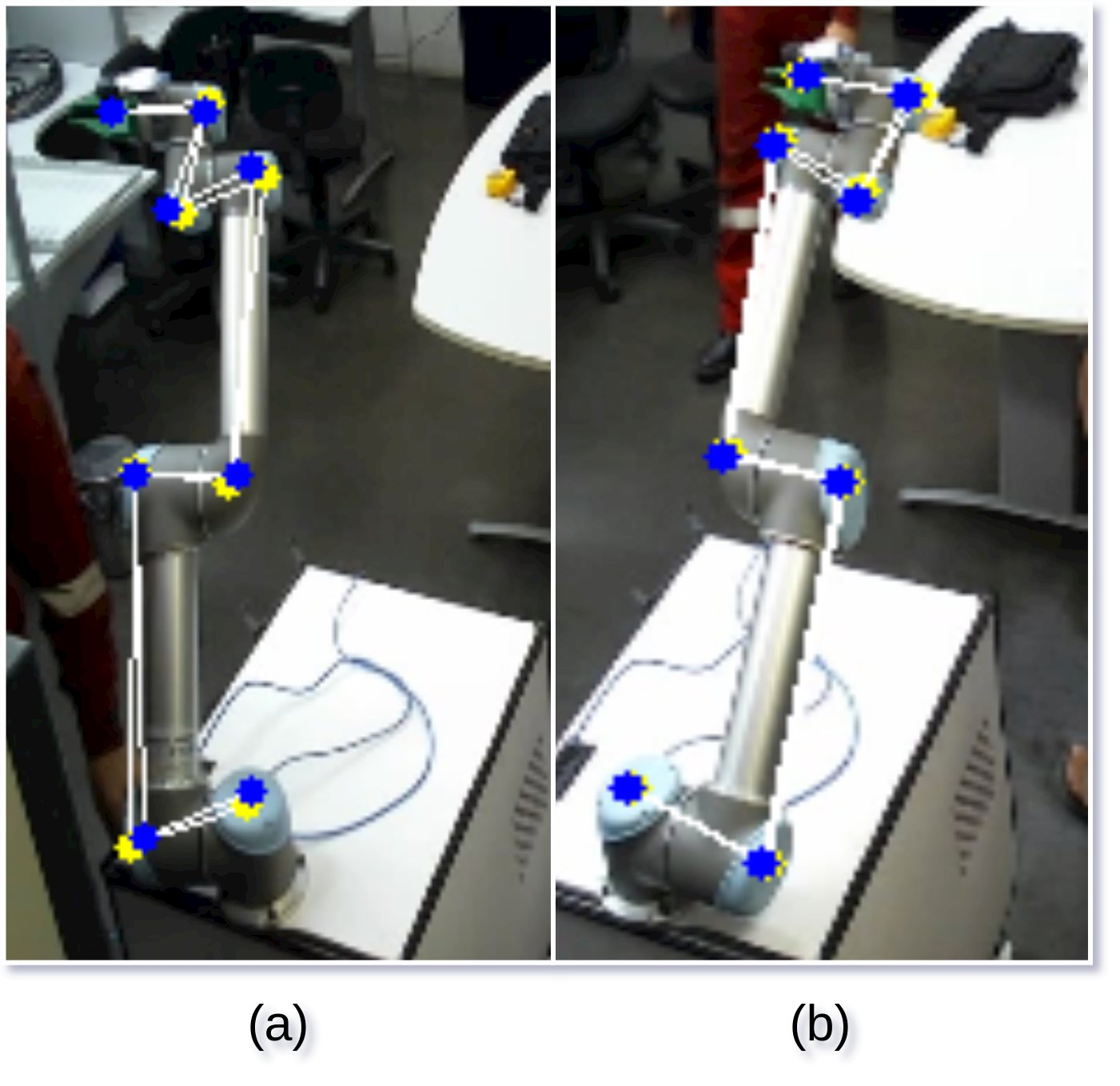}
\caption{Zoom comparison. Complete CNN versus $70\%$ pruning rate.}
\label{fig:zoom_comp_prune_b}
\end{center}
\end{figure}

Figure \ref{fig:comp_prune_a} compares the outcome before and after the proposed refinements. Sub-figures (a) represents the usage of the full CNN. Understandably, CNN segmentation and keypoints detection results are more precise. After the refinement and clustering, the output of pruning at $70\%$, shown in sub-figure (b), produces almost the same result as a full model after the post-processing. Figure \ref{fig:zoom_comp_prune_b} shows in detail the predicted keypoint difference. 

Figure \ref{fig:zoom_comp_prune_b} shows that the yellow points correspond to the results obtained when using the full CNN, whereas the blue points represent the results with $70\%$ pruning. As it is noticeable, they are very close, and in some cases,  the points are so close that it is impossible to distinguish one from another.

\subsubsection{Constrained Device Setup}

As seen in Table \ref{fig:setup}, the Constrained Device scenario, with a Raspberry Pi 3, is very restricted and has only 1GB of {RAM} and a low-power four-core processor. Related to the pruning itself, it keeps the total number of parameters, but it reduces the total of FLOPS on the quantization process. Table \ref{table:prune_quant_rasp} shows these results.

\begin{table}[ht]
\centering
\begin{tabular}{c|cc|ccc}
\cline{2-3}
\textbf{}                                   & \multicolumn{2}{c|}{\textbf{PCK@0.5}}                  & \multicolumn{2}{c}{\textbf{}}                                                                                                  & \multicolumn{1}{l}{}                                                            \\ \hline
\multicolumn{1}{|c|}{\textbf{Pruning Rate}} & \multicolumn{1}{c|}{\textbf{Average}} & \textbf{STD}   & \multicolumn{1}{c|}{\textbf{FPS}}    & \multicolumn{1}{c|}{\textbf{\begin{tabular}[c]{@{}c@{}}Inference \\ Time\end{tabular}}} & \multicolumn{1}{c|}{\textbf{\begin{tabular}[c]{@{}c@{}}Size\\ MB\end{tabular}}} \\ \hline
\multicolumn{1}{|c|}{\textbf{10}}           & \multicolumn{1}{c|}{0.998}            & 0.004          & \multicolumn{1}{c|}{2.860}           & \multicolumn{1}{c|}{0.002}                                                              & \multicolumn{1}{c|}{0.437}                                                      \\ \hline
\multicolumn{1}{|c|}{\textbf{20}}           & \multicolumn{1}{c|}{0.978}            & 0.027          & \multicolumn{1}{c|}{3.098}           & \multicolumn{1}{c|}{0.001}                                                              & \multicolumn{1}{c|}{0.351}                                                      \\ \hline
\multicolumn{1}{|c|}{\textbf{30}}           & \multicolumn{1}{c|}{0.975}            & 0.036          & \multicolumn{1}{c|}{3.814}           & \multicolumn{1}{c|}{0.001}                                                              & \multicolumn{1}{c|}{0.272}                                                      \\ \hline
\multicolumn{1}{|c|}{\textbf{40}}           & \multicolumn{1}{c|}{0.990}            & 0.004          & \multicolumn{1}{c|}{4.299}           & \multicolumn{1}{c|}{0.000}                                                              & \multicolumn{1}{c|}{0.206}                                                      \\ \hline
\multicolumn{1}{|c|}{\textbf{50}}           & \multicolumn{1}{c|}{0.982}            & 0.007          & \multicolumn{1}{c|}{6.684}           & \multicolumn{1}{c|}{0.001}                                                              & \multicolumn{1}{c|}{0.143}                                                      \\ \hline
\multicolumn{1}{|c|}{\textbf{60}}           & \multicolumn{1}{c|}{0.974}            & 0.012          & \multicolumn{1}{c|}{7.764}           & \multicolumn{1}{c|}{0.000}                                                              & \multicolumn{1}{c|}{0.101}                                                      \\ \hline
\multicolumn{1}{|c|}{\textbf{70}}           & \multicolumn{1}{c|}{\textbf{0.969}}   & \textbf{0.009} & \multicolumn{1}{c|}{\textbf{10.040}} & \multicolumn{1}{c|}{\textbf{0.001}}                                                     & \multicolumn{1}{c|}{\textbf{0.063}}                                             \\ \hline
\multicolumn{1}{|c|}{\textbf{80}}           & \multicolumn{1}{c|}{\textbf{0.911}}   & \textbf{0.051} & \multicolumn{1}{c|}{\textbf{15.873}} & \multicolumn{1}{c|}{\textbf{0.001}}                                                     & \multicolumn{1}{c|}{\textbf{0.034}}                                             \\ \hline
\multicolumn{1}{|c|}{\textbf{90}}           & \multicolumn{1}{c|}{0.087}            & 0.092          & \multicolumn{1}{c|}{27.778}          & \multicolumn{1}{c|}{0.002}                                                              & \multicolumn{1}{c|}{0.016}                                                      \\ \hline
\end{tabular}
\caption{Performance Results for Pruning and Quantization over a RaspberryPi 3.}
\label{table:prune_quant_rasp}
\end{table}

\begin{figure}[htp]
\begin{center}
\includegraphics[width=0.7\linewidth]{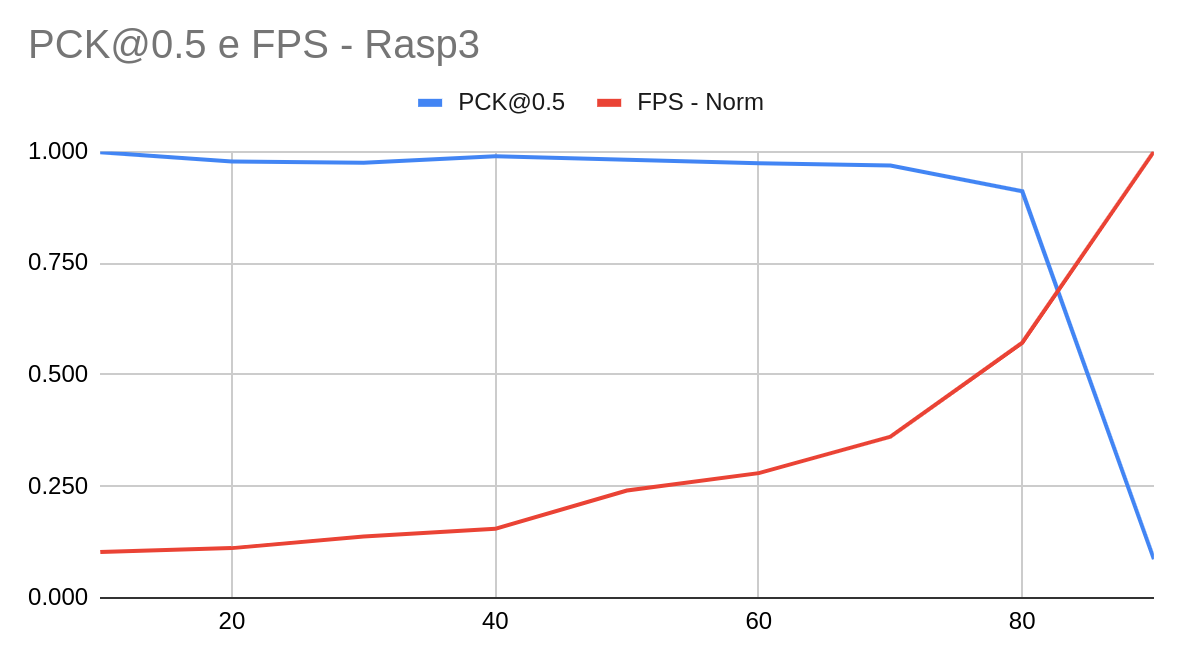}
\caption{Raspberry Pi3 performance curve. PCK versus FPS.}
\label{fig:arm_pck}
\end{center}
\end{figure}

Table \ref{table:prune_quant_rasp} shows that the Raspberry could process 10 FPS at a $70\%$ prune rate while keeping a PCK above $0.96$. Another important metric is the disc storage requirement, where at $70\%$, we only require 0.063 MB. Figure \ref{fig:arm_pck} shows the corresponding performance curve. 

\begin{figure}[htp]
\begin{center}
\includegraphics[width=0.5\linewidth]{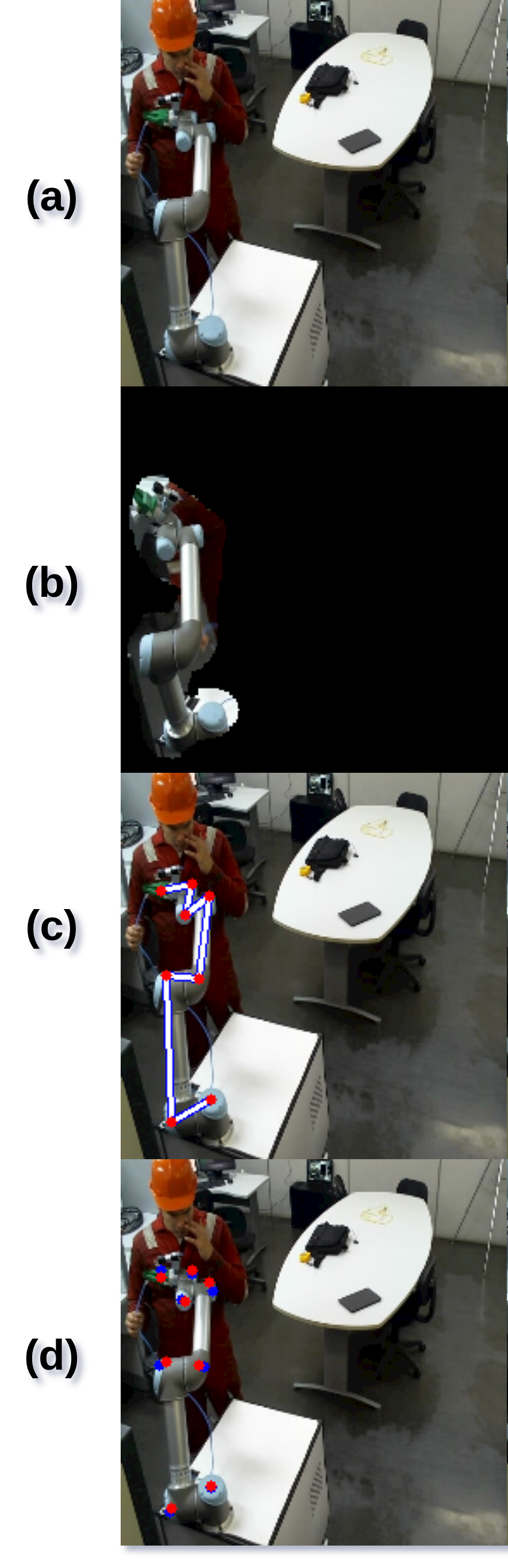}
\caption{The full process of keypoint detection. (a) Input image, (b) segmented arm, (c) and generated keypoints, and (d) a comparison between the real and predicted keypoints.}
\label{fig:arm_seg_pred}
\end{center}
\end{figure}

According to Figure \ref{fig:arm_pck}, we maintain the same behavior as when using the full CNN to a $70\%$ pruning rate. A performance loss is observed with the $80\%$ pruning rate. As a final step, we added a qualitative evaluation of keypoint detection, see Figure \ref{fig:arm_seg_pred} for a view of those samples. 

Figure \ref{fig:arm_seg_pred}(d) presents some promising results:  blue depicts the real keypoint, and red the ones. Observe that it is difficult to even distinguish between the produced real and predicted keypoints in some cases.

\section{Conclusion}
\label{sec:conclusion}

The pose estimation process is usually heavy-duty and often demands robust systems with high computational power. Few approaches deal with pose estimation while trying to reduce a type of demand. In this work, besides proposing a new CNN for pose detection, we also aim to improve the overall response time of the model and its disk storage demands.   

We first generated a new CNN architecture for pose estimation (FCN-Pose) based on segmentation to achieve this goal. Then we proceed to the CNN compression, with pruning and quantization. We perform a combination of both techniques, and evaluate his results with considerable results. Additionally, we evaluate our proposed solution in a real constrained device, expanding the overall understatement on application, deploys, and usability of those solutions.  
 
The FCN-Pose reached a PCK above $0.99\%$ without pruning and quantization, meaning that we keep a decision region close enough in $99\%$ of the cases even with our error, while the prune at $70\%$ only drops $1\%$ in PCK with $0.98\%$. We jump from 11.71 FPS to 41.90 FPS on the Desktop setup with $70\%$ pruning. And finally, we the disc storage requirement from 1.7 MB to just 0.248 MB. 

On the Constrained Device Setup, with the FCN-Pose, pruned and quantized, we could get a PCK of $0.96\%$ on $70\%$, allowing us to run the FCN-Pose, from 2.8 FPS to 10.04 FPS, requiring only 0.063 MB on disc storage. 

It is valid to notice that all those values of pruned were acquired on the performance curve, but may a PCK smaller can be applicable in some cases. For instance, on the Constrained Device, a prune of $80\%$ returns a PCK of $0.91\%$ but increases our FPS to 15.87. 

Finally, for future works, our solution is applicable on constrained devices, but we could deploy them in distributed networks, as nodes, for instance. Another important point is that our solution was designed for robot pose estimation with the HRI. In the current stage, we can improve this tool by adding the human pose estimation as a factor and trying to make a decision based on both inputs.

\bibliographystyle{IEEEtran}
\bibliography{bare_jrnl.bib}

\end{document}